\documentclass[10pt, a4paper, conference , twocolumn]{IEEEtran}
\usepackage[utf8]{inputenc}
\usepackage[T1]{fontenc}
\usepackage{amsmath}
\usepackage{graphicx}
\usepackage{url}
\usepackage{subcaption}

\usepackage{xcolor}

\usepackage{fancyhdr}
\usepackage[a4paper, hmargin=13mm, vmargin={27mm,45mm}, headsep=4mm]{geometry}

\renewcommand{\headrulewidth}{0pt}
\begin{document}
\title{Face Recognition using Compressive Sensing\\\large{(Student Paper)}}

\author{
% First author - Prvi autor
\IEEEauthorblockN{S. Kovačević }
\IEEEauthorblockA{
University of Montenegro\\
Podgorica, Montenegro \\
Email: skovacevic96@outlook.com}
% Second author - Drugi autor
\and
\IEEEauthorblockN{V. Đaletić }
\IEEEauthorblockA{
University of Montenegro\\
Podgorica, Montenegro \\
Email: djaleticv@gmail.com}
\and
\IEEEauthorblockN{J. Vuković }
\IEEEauthorblockA{
University of Montenegro\\
Podgorica, Montenegro \\
Email: jelenavukovic51@gmail.com}
% Other authors - Ostali autori
}
\maketitle \thispagestyle{fancy} % Do not change this - ne mijenjajte ovu liniju

\begin{abstract}
This paper deals with the Compressive Sensing implementation in the Face Recognition problem. Compressive Sensing is new approach in signal processing with a single goal to recover signal from small set of available samples. Compressive Sensing finds its usage in many real applications as it lowers the memory demand and acquisition time, and therefore allows dealing with huge data in the fastest manner. In this paper, the undersampled signal is recovered using the algorithm based on Total Variation minimization. The theory is verified with an experimental results using different percentage of signal samples. 
\end{abstract}


\section{Introduction}
 Face Recognition is a technology capable of identifying or verifying a person from either a digital image or from a video source. Primarily seen as a computer application, it is now used on mobile platforms and many other forms of technology. Face Recognition is used alongside other biometrics systems, such as fingerprint or eye recognition. However, it is important to note that its accuracy is usually near the other biometrics but its non-invasive nature makes it a favorable choice. This practicality of Face Recognition is a state of prolonged public dispute due to its more recent abuse by governmental as well as non-governmental usage which ranges from commercial tools to security.
\par
Increase of digitalization allowed Face Recognition technologies to flourish and by Moore's law terms, the error rate is expected to decrease by one-half every two years. Many different approaches are used in the process of extracting the face features and even though they can be fundamentally different, a lot of solutions are based on mixing these approaches in order to maximize the accuracy. Principal component analysis is used in order to decrease dimensionality of the features and to speed up the process of image processing. Some security systems use infrared illuminators to acquire images which results in faces lacking mustaches, beards, hairstyle and glasses which are often used to trick recognition systems. Much of the information about Face Recognition is, however, not easily accessed after 9/11 as security threats made their nature completely covert and they are not publicized. 
\par
Main advantage of a facial recognition system is its ability to mass identify a lot of persons without their cooperation which is widely used in security zones and public places but on social networks as well. The individual could be easily identified in the crowd.  Face Recognition is effective at full frontal faces and at faces that are about 20 degrees off but the profile images are usually non usable. Even the facial expressions like big smiles can decrease its effectiveness and that is the reason Face Recognition systems are still not able to beat human benchmark. Government issued IDs that allow only neutral facial expressions and full fronting faces. Low resolution pictures often do not contain enough features to do valid identification which is a problem as increasing resolution slows down computer systems. Replicated faces are able to trick modern Face Recognition which is a huge concern. Social networks are about to change this as their databases are growing everyday, giving more data to the algorithms and eventually making them more successful. However, more data means more memory and bigger pictures means more time spent on processing. As Compressive Sensing allows reconstruction of the signal using small number of data, the possibility to apply this novel sampling and recovery method in Face Recognition is tested in this paper.
\par
Privacy violations are increasing concern as security systems implemented in Western countries and recently China completely reshaped our day-to-day privacy. An individual is not able to prevent face recognition without completely covering the face which can, as some fear, lead to total surveillance society. Social networks use face recognition not just to recognize a face but to unearth entire personal data by identifying photos featuring the individual, blog posts, internet behavior which can lead to a patern that is completely privacy invasive. 

\section{Compressive sensing}
Sampling theorem also know as Nyquist-Shannon theorem states that the signal can be reconstructed if sampling rate is at least twice the highest frequency of that signal. 
When we talk about photos, videos and other signals where we have a large number of samples which can make whole process of signal acquisition and analysis very complex and also expensive. This brings us to different approach in signal processing which is know as the compressive sensing, compressive sampling or sparse sampling (CS). With compressive sensing we can process signals and provide high quality signal reconstruction using significantly lower number of samples compared with the Nyquist-Shannon theorem. The reconstruction is based on using optimization algorithms. The reconstruction is possible if some conditions are satisfied. Namely, samples should be acquired in random manner and signal should have sparse representation in certain transformation domain (depending on type of signal, sparsity domain can be time, frequency, time-frequency, wavelet domain, etc.). Sparse signal keeps important information in small number of non-zero coefficients in the sparsity domain. On the other hand, when considering 2D data, we can't expect that image have most zero value pixels. Therefore, images are not strictly sparse. The suitable sparsity domain for images is the Discrete cosine transform (DCT) domain. In this domain we can consider image as nearly sparse.
\par
Let us shortly introduce the basis mathematics behind the CS. If we have signal $\boldsymbol{x}$ whose length is $ N$, this signal can be represented as:
\begin{equation}
\label{eq:x}
\boldsymbol{x} =\boldsymbol{\Psi} \boldsymbol{v}
\end{equation}
where $\boldsymbol{\Psi}$ is transform matrix (in our case, it's DCT transform matrix), and columns of that matrix are basis vectors. The vector $\boldsymbol{v}$ represents signal in transform domain.
The mathematical determination of signal reconstruction by small number of samples is not a simple problem. It is assumed that process of collecting measurement data is linear and it is represented as a linear equation system: 
\begin{equation}
\label{eq:A}
\boldsymbol{A}\boldsymbol{x} = \boldsymbol{y}
\end{equation}
 The matrix $\boldsymbol{A}$ is the submatrix of the  $\boldsymbol{\Psi}$ matrix, and it's formed with rows or columns which are selected randomly from $\boldsymbol{\Psi}$ matrix. Vector $\boldsymbol{y}$ is vector of CS measurements and this vector dimensions are $M\times 1$. 
 Equation \ref{eq:A} has infinite number of possible solutions. In order to  find the  sparsest solution which will lead to the reconstruction of the signal $\boldsymbol{x}$, the optimization algorithms are used. In this paper we will describe total variation minimization, which is used for the reconstruction of the undersampled face images.
Total variation algorithm is based on minimization of the image gradient, that is sparse in the transform domain. Total variation of the signal (image) $\boldsymbol{x}$ is defined as sum of the magnitudes of discrete gradient: 
\begin{equation}
\label{eq:TV}
TV(\boldsymbol{x}) = \sum_{i,j}^{}\sqrt{D_{h_{ij}}^{2}x + D_{v_{ij}}^{2}x} = \sum_{i j}^{}\left \|D_{i,j}\boldsymbol{x}\right \|_{2}
\end{equation}
where $D_{i j}\boldsymbol{x}$ represents discrete gradient of the image $\boldsymbol{x}$. 
\begin{equation}
\label{eq:Dh}
D_{h_{ij}}\boldsymbol{x} =\left \{\begin{matrix} x_{i+1, j} - x_{i j} & i < n\\  0 & i = n  \end{matrix}\right.
\end{equation}
\begin{equation}
\label{eq:Dv}
 D_{v_{i,j}}\boldsymbol{x} = \left\{\begin{matrix} x_{i, j+1} - x_{i j} & j < n\\  0 & j = n  \end{matrix}\right.
\end{equation}
\begin{equation}
\label{eq:D}
D_{i j}\boldsymbol{x} = \begin{pmatrix} D_{h_{ij}}\boldsymbol{x}\\  D_{v_{i j}}\boldsymbol{x} \end{pmatrix} 
\end{equation}
where $i$ is for row and $j$ is for column of $n\times n$ image size. 
Signal reconstruction based on CS is less expensive and simpler and has a wide variety of applications. One of these applications is facial recognition which will be described in detail in next chapter.

\section{Face Recognition} 
Given the dataset of facial images of people that are ought to be recognized, when an input image is presented Face Recognition algorithm matches the face in the input image to a person in the dataset [18]. After acquiring face gallery, a process of feature extraction is then done in order to store features of interest in a feature vector. After that, machine learning algorithm is used in order to learn to discriminate between persons in the dataset. Output of the learning process is called classifier that will be used for face recognition. 
\begin{figure}[h!]
	\includegraphics[width=\linewidth]{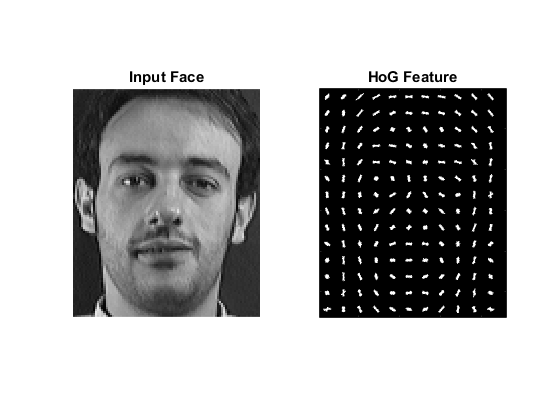}
	\caption{HOG features of a random face in the dataset.}
	\label{fig:hog}
\end{figure}
\par Every input frame is scanned and after a face is detected it is cropped and then resized in order to match the data in the dataset. After extracting the face, we repeat the process of feature extraction. Those features are compared with classifier which gives us a label of the person matched with input face. It is very important to split the dataset so the images trained on are not used for testing. Here 80\% to 20\% partition is used. After the partitioning, Histogram of oriented gradients (HOG) features are extracted from each image. These features encode the edge information and directionality of edges which gives us a fair sensing of an object or face in this paper.Although much better features could be obtained by using deep neural networks, thus allowing us to hope for better results, it is important to understand that getting a perfect face recognition algorithm was not the goal of this paper. Fitting multiclass model is then done by using 45 binary support vector machine (SVM) models using the one-versus-one coding design. 
\section{Experimental Results}
The experiments were done on picture gallery with 40 different persons, each contributing with 10 images which were purposely taken from different angles in order to generalize the problem. Size of every image is $92\times112$ pixels and they are grayscaled as color is not the feature used by the algorithm. Total of 256 grey levels per pixel were used. For some subjects, the images were taken at different times, varying the lighting, facial expressions (open / closed eyes, smiling / not smiling) and facial details (glasses / no glasses). All the images were taken against a dark homogeneous background with the subjects in an upright, frontal position (with tolerance for some side movement). [17]
\begin{figure}[h!]
	\includegraphics[width=\linewidth]{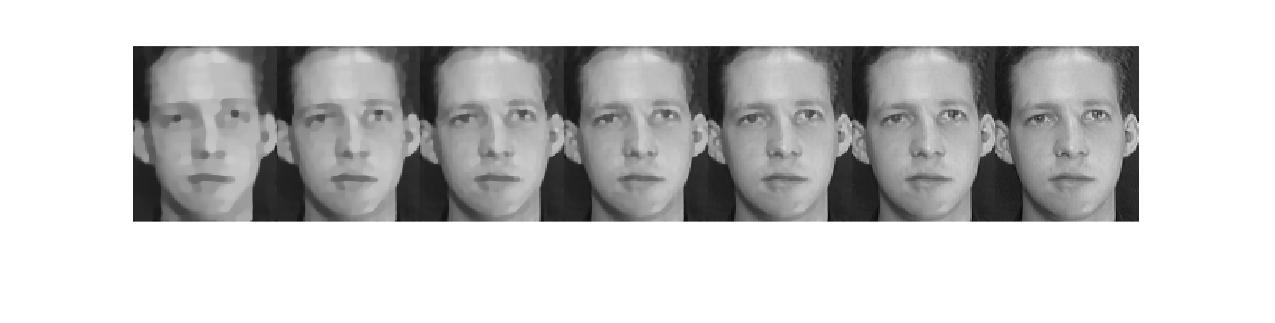}
	\caption{a)5\% of high frequency coefficients retained b) 10\% c) 15\% d) 20\% e)30\% f) 45\% g) 100\% for first persion in dataset}
	\label{fig:coefficients}
\end{figure}
\par
\begin{figure}[h!]
	\includegraphics[width=\linewidth]{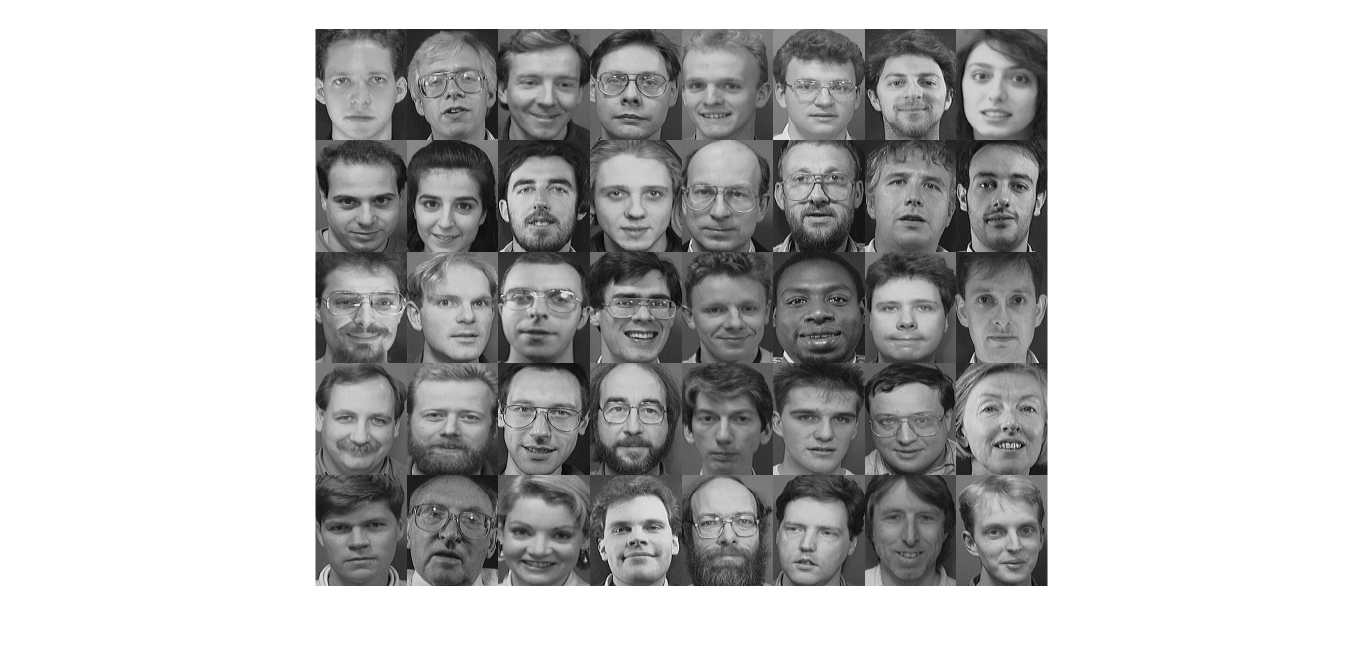}
	\caption{All faces in the database.}
	\label{fig:faces}
\end{figure}

 Figure \ref{fig:faces} shows all the faces in the database. Beware that this is the only 1 of 10 possible angles. The database is split so that 80\% is used to train Face Recognition algorithm and other 20\% are used for testing. Compressive Sensing is done only on the test set and the training is done on the original images. This property of Face Recognition sped these operations a lot. It is important to stress out that 1\% of the low frequency coefficients were used in every CS experiment, and the rest of the coefficients are taken from middle and high frequency image regions. On Figure \ref{fig:coefficients} you can see the images of person 1 after the reconstruction whilst the far right image is original one. 
Even after retaining only 5\% of the image it looks easily recognizable even though the overall quality is squashed. 

 \par Our visual intuition can be easily proven by calculating Peak signal-to-noise ratio (PSNR). Even after reducing image i.e. considering only 5\% of image coefficients as available, we still can obtain satisfactory PSNR, whose value is around 24 dB. Images with 30\% or more coefficients available have larger PSNR, which is visually confirmed as well. Images with low percentage of available coefficients look like aquarell and in some extreme cases it would be almost impossible for human to match the faces but the algorithm was still able to do the recognition. This proves that HOG features are resilient to noise.
\begin{figure}[h!]
	\includegraphics[width=\linewidth]{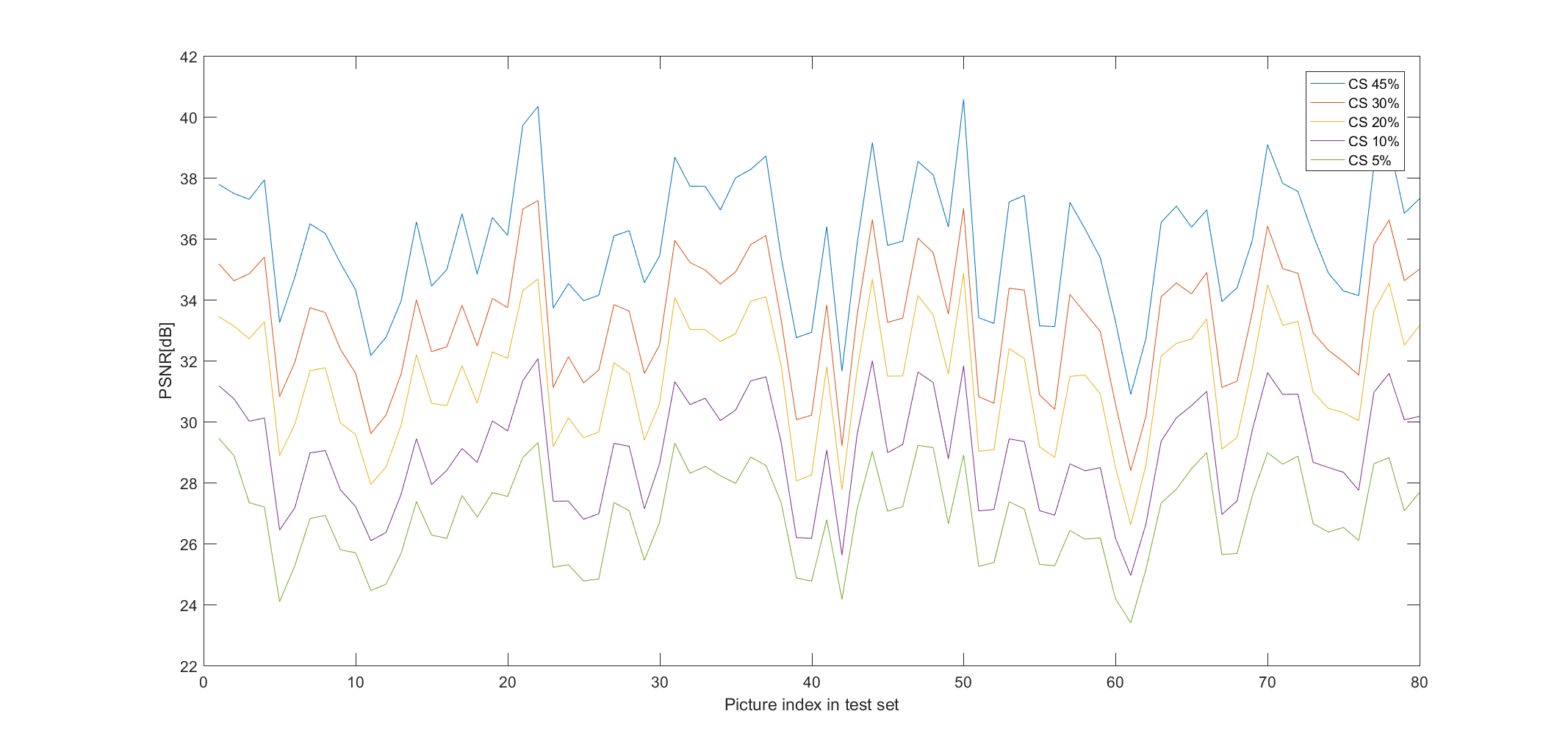}
	\caption{PSNR in relation to percentage of details retained.}
	\label{fig:PSNR}
\end{figure}

\par
\begin{figure}[h!]
	\includegraphics[width=\linewidth]{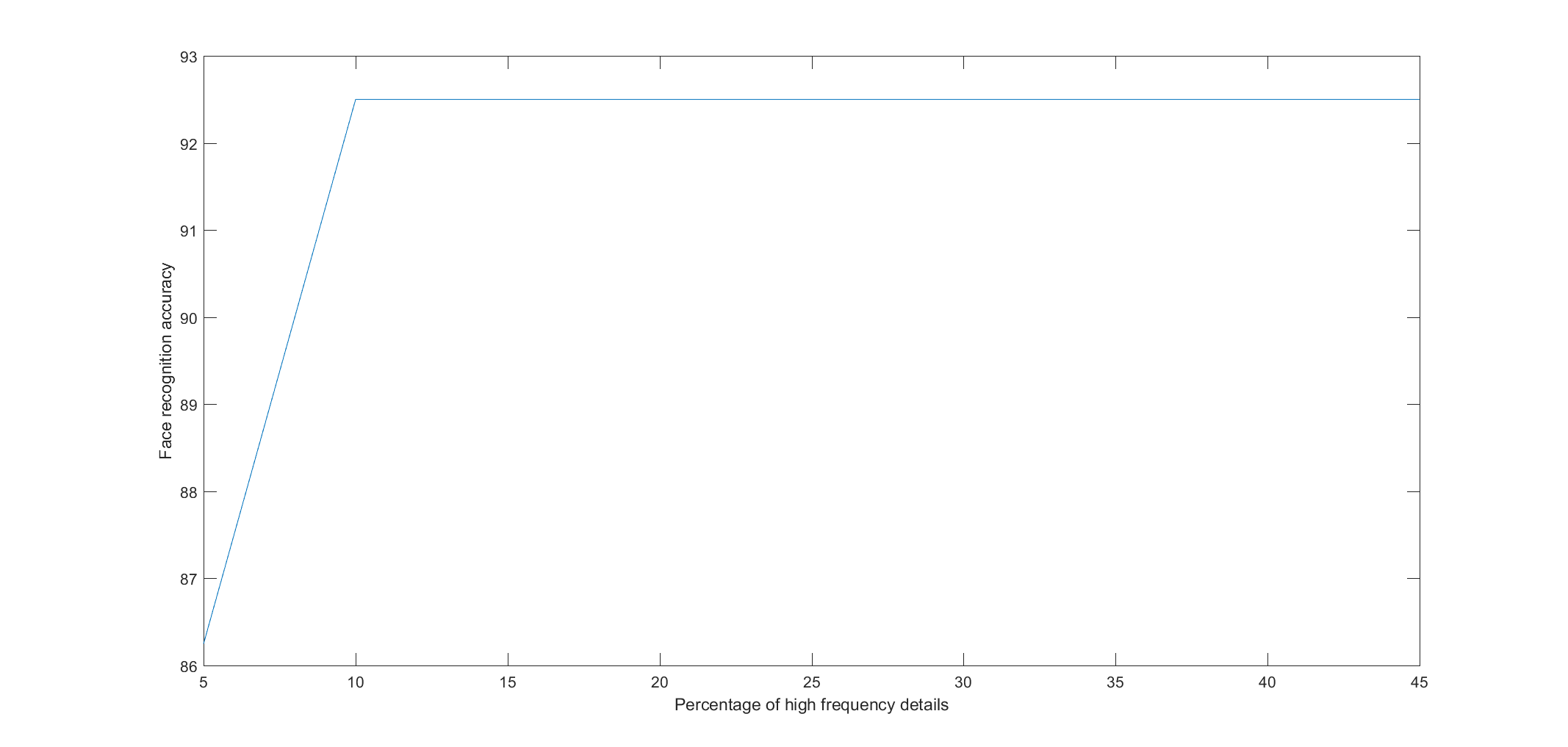}
	\caption{Face Recognition accuracy.}
	\label{fig:accuracy}
\end{figure}

 The most important result of this paper is the accuracy of Face Recognition on test data which was reconstructed after CS. These results are satisfying.
 Even after reducing image to only 10\% of its details, we get 92.5\% (as shown on figure \ref{fig:accuracy}) accuracy which matches the results of the algorithm used on the original test set. Success of Face Recognition on 5\% CS data is good overall, but from a sampling perspective the difference is not that big and the advice would be to sacrifice the acquisition time a bit in order to improve accuracy so at least 10\% CS images should be used. The exact percentage is expected to be even higher as practical conditions could differ from ones used in this experiment.
\begin{figure}[h!]
	\begin{center}
	\includegraphics[scale = 0.4]{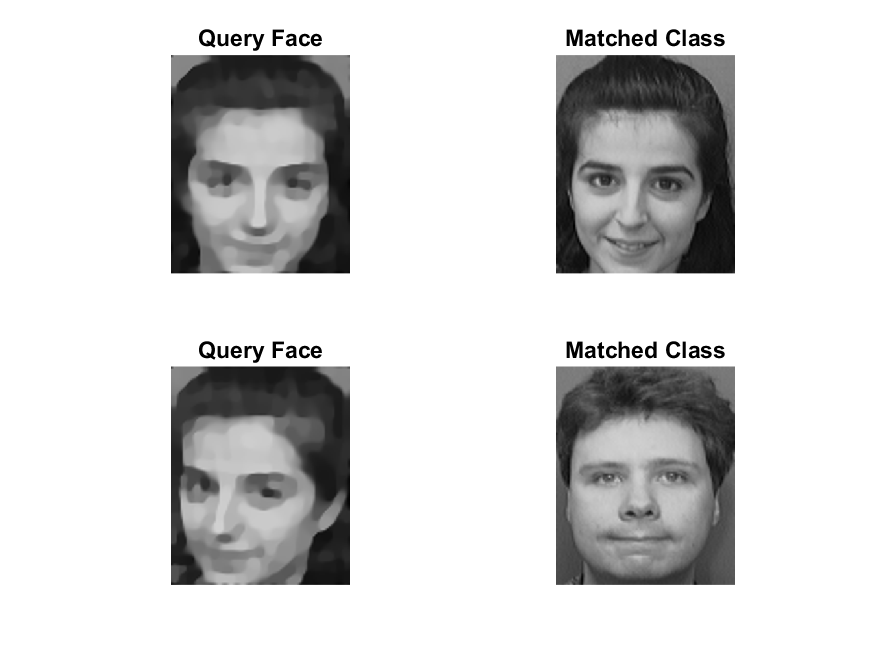}
	\end{center}
	\caption{Query example, 5\% of coefficients used.}
	\label{fig:query}
\end{figure}
\par
\begin{figure}[h!]
	\includegraphics[width = \linewidth]{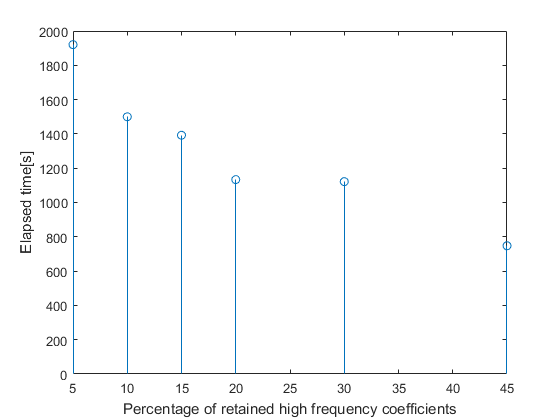}
	\caption{Time spent on Compressive Sensing of test data.}
	\label{fig:timespent}
\end{figure}

 On Figure \ref{fig:query} we can see examples of both correct match and a missmatch where two photos of the same person were queried. In the case of the missmatch, error seems more drastic as queried face and matched one have nothing in common, not even their gender. 
Time spent on reconstruction of images grows as percentage of retained coefficients decrease. The time shown on the Figure \ref{fig:timespent} is spent on the entire test data or 80 images in total. 

\section{Conclusion}
Implementation of CS in Face Recognition is presented in the paper. While using simple Face Recognition algorithm and a rather modest Total Variation technique for CS it can be concluded that it is possible to compress most of the data and still preserve the accuracy. We measured reconstruction quality with PSNR which backs our visual intuition and arithmetic calculations. Testing this approach on the much bigger database is the next step which will hopefully bring more confidence to our results.


\begin{thebibliography}{99}
\bibitem{a1}S. Stanković, I. Orović, and E. Sejdić, Multimedia Signals and Systems: Basic and Advance Algorithms for Signal Processing,. Springer-Verlag, New York, 2015 
\bibitem{a2}J. A. Tropp. “Greed is good: Algorithmic results for sparse approximation,” IEEE Transactions on
Information Theory, vol. 50, no. 10, 2004, pp.2231-2242. 
\bibitem{a3}N. Lekić, A. Draganić, I. Orović, S. Stanković, and V. Papić, “Iris print extracting from reduced and scrambled set of pixels,” Second International Balkan Conference on Communications and Networking BalkanCom 2018, Podgorica, Montenegro, June 6-8, 2018 
\bibitem{a4}N. Lekić, M. Lakičević, I. Orović, and S. Stanković, “Adaptive gradient-based analog hardware architecture for 2D under-sampled signals reconstruction,” Microprocessors and Microsystems, Volume 62, October 2018, Pages 72-78 
\bibitem{a5}E. Sejdic, "Time-Frequency Compressive Sensing", in Time-Frequency Signal Analysis and Processing, ed. B. Boashash, Academic Press, Nov. 2015. pp.424-429.
\bibitem{a6}G. Pope, “Compressive Sensing: a Summary of Reconstruction Algorithms,” Eidgenossische Technische Hochschule, Zurich, Switzerland, 2008
\bibitem{a7}A. Draganić, I. Orović, S. Stanković, X. Li, and Z. Wang, “An approach to classification and under-sampling of the interfering wireless signals,” Microprocessors and Microsystems, Volume 51, June 2017, pp. 106–113.
\bibitem{a8}LJ. Stanković, E. Sejdić, S. Stanković, M. Daković, and I. Orović, “A Tutorial on Sparse Signal Reconstruction and its Applications in Signal Processing,” Circuits, Systems $\And$ Signal Processing, published online 01. Aug. 2018 
\bibitem{a9}E. J. Candes and M. B. Wakin, “An introduction to compressive sampling,” IEEE Signal Process. Mag., vol. 25, no. 2, pp. 21–30, 2008.
\bibitem{a10}L. Stankovic, I. Orovic, S. Stankovic, and M. Amin, “Compressive Sensing Based Separation of Non-Stationary and Stationary Signals Overlapping in Time-Frequency,” IEEE Transactions on Signal Processing, vol. 61, no. 18, 2013, pp. 4562-4572.
\bibitem{a11}S. Stanković, and I. Orović, “An Approach to 2D Signals Recovering in Compressive Sensing Context,” Circuits, Systems and Signal Processing, April 2017, Volume 36, Issue 4, pp. 1700–1713 
\bibitem{a12}M. Fornsaier, H. Rauhut, “Iterative thresholding algorithms,” Applied and Computational Harmonic Analysis, vol. 25, no. 2, 2008, pp. 187- 208. 
\bibitem{a13}I. Orovic, S. Stankovic, and T. Thayaparan, “Time-Frequency Based Instantaneous Frequency Estimation of Sparse Signals from an Incomplete Set of Samples,” IET Signal Processing, Special issue
on Compressive Sensing and Robust Transforms, vol. 8, no. 3, 2014, pp. 239-245 
\bibitem{a14}I. Orovic, and S. Stankovic, “Improved Higher Order Robust Distributions based on Compressive Sensing Reconstruction,” IET Signal Processing, vol. 8, no. 7, 2014, pp. 738-748
\bibitem{a15} J. Romberg, “Imaging via Compressive Sampling,” IEEE Signal Processing Magazine, March 2008.
\bibitem{a16}J. Music, T. Marasovic, V. Papic, I. Orovic, and S. Stankovic, “Performance of compressive sensing
image reconstruction for search and rescue,” IEEE Geoscience and Remote Sensing Letters, vol. 13, no. 11, 2016, pp. 1739-1743 
\bibitem{a17} The Database of Faces, Cambridge University Computer Laboratory\\
\url{https://www.cl.cam.ac.uk/research/dtg/attarchive/facedatabase.html}
\bibitem{a18} MATLAB Face Recognition \\
\url{https://www.mathworks.com/discovery/face-recognition.html}

\end{thebibliography}
\end{document}